\documentclass[sigconf]{acmart}
\renewcommand\footnotetextcopyrightpermission[1]{} 
\usepackage{booktabs} 
\usepackage{todonotes}
\usepackage{soul}
\usepackage{pifont}
\usepackage{rotating}
\usepackage{multirow}

\newcommand{\cmark}{\ding{51}}%
\newcommand{\xmark}{\ding{55}}%
\newcommand{\vrt}[1]{\begin{turn}{90}#1\end{turn}}

\makeatletter
\newcommand\footnoteref[1]{\protected@xdef\@thefnmark{\ref{#1}}\@footnotemark}
\makeatother

\begin{document}
\title{Learning Linear Feature Space Transformations in~Symbolic~Regression}

\author{Jan \v{Z}egklitz}
\orcid{http://orcid.org/0000-0003-3302-6779}
\affiliation{%
    \institution{Czech Technical University in Prague\\
        Czech Institute of Informatics, Robotics and Cybernetics}
    \streetaddress{Jugosl\'{a}vsk\'{y}ch partyz\'{a}n\r{u} 1580/3}
    \city{Prague 6} 
    \state{Czech Republic} 
    \postcode{160 00}
}
\email{zegkljan@fel.cvut.cz}

\author{Petr Po\v{s}\'{i}k}
\orcid{http://orcid.org/0000-0002-9694-3673}
\affiliation{%
    \institution{Czech Technical University in Prague\\
        Faculty of Electrical Engineering}
    \streetaddress{Technick\'{a} 1902/2}
    \city{Prague 6} 
    \state{Czech Republic} 
    \postcode{166 27}
}
\email{petr.posik@fel.cvut.cz}


\begin{abstract}
We propose a new type of leaf node for use in Symbolic Regression (SR) that performs linear combinations of feature variables (LCF).
These nodes can be handled in three different modes -- an unsynchronized mode, where all LCFs are free to change on their own, a synchronized mode, where LCFs are sorted into groups in which they are forced to be identical throughout the whole individual, and a globally synchronized mode, which is similar to the previous mode but the grouping is done across the whole population.
We also present two methods of evolving the weights of the LCFs -- a purely stochastic way via mutation and a gradient-based way based on the backpropagation algorithm known from neural networks -- and also a combination of both.
We experimentally evaluate all configurations of LCFs in Multi-Gene Genetic Programming (MGGP), which was chosen as baseline, on a number of benchmarks.
According to the results, we identified two configurations which increase the performance of the algorithm.
\end{abstract}

%
%
%




\maketitle

\section{Introduction} \label{sec:intro}
Symbolic regression (SR) is an inductive learning task with the goal to find a model in the form of a symbolic mathematical expression that fits the available training data.
SR is a landmark application of Genetic Programming (GP) \cite{Koza1992}.
GP is an evolutionary optimization technique that is inspired by biological evolution to evolve computer programs that perform well in a given task.


Recently, several methods emerged \cite{Searson2015-published,McConaghy2011,Arnaldo2015,Arnaldo2014} that explicitly evolve models in a form of (possibly regularized) ``top-level'' linear combinations of evolved complex features.
Such models can be learned much faster since the evolution does not have to deal with the linear parts.

In some SR tasks, the underlying function could be modeled more easily if we had access to a suitable rotation of the feature space, or to suitable projections of the features.
Such transformations can be achieved by linear combinations of the features of the problem.
Such linear combinations are already available in virtually any SR system that allows for numeric constants.
These are usually tuned by mutation and the linear combinations must be constructed via structural manipulation operators.
In this article we explore the possibility of using explicit linear combinations of features at the bottom of the evolved expression trees.
These are added to the original features and can then be non-linarly combined by evolution.

We have chosen Multi-Gene Genetic Programming (MGGP) \cite{Hinchliffe1996,Searson2015-published} as the base algorithm for the research as it is very close to regular GP but uses top-level linear combinations to speed up the search.

The rest of the article is organized as follows: in Section \ref{sec:mggp} we briefly review MGGP as it is the foundation of our work.
Section \ref{sec:descr} we describe the linear combinations of features and how do they work.
Section \ref{sec:experiments} is dedicated to the description of the experimental evaluation of our proposals.
Section \ref{sec:results} presents the results and discusses the implications.
Section \ref{sec:conclusion} concludes the paper and provides suggestions for future work.

\section{Multi-Gene Genetic Programming} \label{sec:mggp}
MGGP \cite{Hinchliffe1996,Searson2010,Searson2015-published} is a tree-based genetic programming algorithm utilizing multiple linear regression.
The main idea behind MGGP is that each individual is composed of multiple independent expression trees, called genes, which are put together by a linear combination to form a single final expression.
The parameters of this top-level linear combination are computed using multiple linear regression where each gene acts as an independent feature.

In this article we base upon a particular implementation of MGGP -- GPTIPS2 \cite{gptips-source}\footnote{
    We didn't use that implementation but re-implemented it.
}.
This particular instance of MGGP uses two crossover operators: (i) high-level crossover that, according to a probability which is a parameter of the algorithm, selects genes from two parents and swaps them between those parents, throwing off the genes that would exceed the upper limit on the number of genes; (ii) low-level crossover which is a classical Koza-style subtree crossover.

Also, there are two mutation operators: (i) subtree mutation which is a classical Koza-style subtree mutation; (ii) constant mutation which mutates the numerical values of leaves representing constants by adding a normally distributed random number.

Both the crossover and mutation operators are chosen stochastically.
\section{Linear Combinations of Features} \label{sec:descr}
We introduce a new type of leaf node -- a Linear Combination of Features or LCF for short.
This type of node is similar to a leaf node representing a variable, or feature. However, while an ordinary feature-node evaluates simply to the value of that feature, a LCF node evaluates to a linear combination of all the features present in the solved problem.
Mathematically, the node implements the following function
\begin{equation}
lcf(\mathbf{x}) = a + \mathbf{bx}
\end{equation}
where $\mathbf{x}$ is the vector of feature values and $a$ and the vector $\mathbf{b}$ are constants.

The LCFs effectively perform affine transformations of the feature space and we argue that they can provide more effective tools to deal with e.g. rotated functions and, in general, provide more flexibility to the GP algorithm.

\paragraph{Terminology} In the rest of the paper we will refer to the additive constant $a$ and multiplicative coefficients $\mathbf{b}$ simply as to \emph{weights}.

\subsection{Initialization of LCFs}
LCF nodes have two or more weights (depending on the dimensionality of the problem) that have to be somehow initialized at the start of the algorithm.

The initialization method is based on the idea that, at the start, there is no feature space transformation happening.
This means that each LCF is initialized such that the additive weight and all multiplicative weights except for one are set to zero.
The only non-zero multiplicative weight is set to one\footnote{
    \label{footnote:random}
    We also tried random initialization -- all weights sampled randomly -- but preliminary experiments showed no benefit over the described init. method.
}.

\subsection{Tuning of LCF weights}
In order to be of any use, the weights of LCF nodes must be modified during the evolution.
We present two such methods.

The first approach, probably the simplest possible, is \emph{weights mutation}, i.e. a dedicated mutation operator.
We use an approach similar to one used for tuning constants in leaf nodes -- a gaussian mutation.
When weights mutation occurs, a single random LCF node is selected and each weight is offset by a random number from gaussian distribution.

The second approach is \emph{gradient-based tuning}.
Weights mutation is simple but it is not informed as it relies just on the selection strategy to promote good mutations.
However, since the structure of the expressions as well as the cost function is known, it is possible to compute the gradient of the expressions w.r.t. the weights.
We use an approach fundamentally identical to the one used in neural networks -- error backpropagation technique\footnote{
    We strictly separate the task of determining the values of the partial derivatives (i.e. the gradient) w.r.t. the parameters and the actual update of the parameters.
    When we use the term ``(error) backpropagation'', we mean only the procedure of determining the values of the partial derivatives and not the update mechanism.
} \cite{Rumelhart1986}.
When the individual partial derivatives are known, any first-order update method can be used to modify the weights to produce more fit expression.

In our setup we use the iRprop$^-$ update mechanism \cite{Igel2000}.
There are two reasons for using this particular method:
\begin{enumerate}
    \item it is very simple to implement yet is very efficient \cite{Igel2003}\footnote{
        Although \cite{Igel2003} shows that iRprop$^+$ is superior to iRprop$^-$, our preliminary experiments have shown that using iRprop$^+$ causes more overfitting and not as good results, hence we decided to use iRprop$^-$.
    }, and
    \item is numerically robust as it operates only with the signs of the partial derivatives rather than their magnitudes.
\end{enumerate}
The second reason is especially important one because in the GP environment there are (generally) no constraints on the inner structure of the expressions.
Due to the finite precision of binary representation of real numbers in computers, this can lead to infinite derivatives even if, mathematically, they are finite.

\subsection{Operation modes of LCFs}
When an LCF node is generated (during initialization or structural mutation) an index is assigned to it.
This index can be any integer between (and including) 1 and the number of the features of the solved problem.
We define three operation modes that differ in how the weights of LCFs are handled in relation to their index.

\subsubsection{Unsynchronized mode}
In the \emph{unsynchronized} mode there are no restrictions whatsoever on the weights.
In other words, any LCF is allowed to have any weights, regardless of its index.

\subsubsection{Synchronized mode} \label{sec:synced}
In this mode, all LCFs with the same index in a particular individual are forced to have the same weights.
This way, all the LCFs in a model form a single affine transformation of the feature space, effectively producing a simpler model.

From a technical point of view, all LCFs are still treated independently.
In order to get the desired behaviour, a special handling is required:
\begin{itemize}
    \item After each backpropagation phase (if tuned by gradient-based approach) the values of partial derivatives for the parameters are summed up inside the index-determined groups.
    This way, all such nodes will be updated in the same way.
    \item If it is detected that a model has, for some of the indexes, two sets of weights (a result of a structural mutation or a crossover with another model), both sets and their arithmetic mean are evaluated and the best performing setting is used.
\end{itemize}

\subsubsection{Globally synchronized mode}
The \emph{globally synchronized} mode is similar to the synchronized mode but the index-based synchronization encompasses the whole population instead of single models.
The motivation behind this mode is that should there truly be a globally suitable transformation of the input space in the data, the models can all work together to find this transformation.

Since there is only a single set of LCFs, using only mutation to tune them makes no sense because there is no population of LCF sets the selection could pick from and hence mutation would be only a random walk.
Therefore we always use gradient-based approach, alone or accompanied by the mutation as a means to help it escape from local optima.
\section{Experimental evaluation} \label{sec:experiments}
In this section we describe the experimental setup for the evaluation of our proposal.

\subsection{Algorithm configurations}
We proposed several ways of how LCFs can work in two aspects of the algorithm: operation mode, and method of tuning the LCF weights.
In the following text we shall use ``codenames'' for each of the algorithm configurations.

Each codename is composed of two letters.
The first letter describes the LCF operation mode and second letter describes how are the LCF weights tuned.
The code letters are described in Table \ref{tab:codenames}.
The baseline, i.e. unmodified MGGP, has none of the two letters and will be referred to as \emph{baseline}, or (in tables) by two dashes, one for each missing letter.
\begin{table}[htb]
    \centering
    \caption{Description of codename letters.}
    \label{tab:codenames}
    \begin{tabular}{ccc}
        \toprule
        \textbf{aspect} & \textbf{code letter} & \textbf{description} \\
        \midrule
        \multirow{3}{*}{operation mode} & U & unsynchronized \\
        & S & synchronized \\
        & G & globally synchronized \\
        \midrule
        \multirow{3}{*}{weights tuning} & M & mutation \\
        & B & backpropagation \\
        & C & all, i.e. both M and B \\
        \bottomrule
    \end{tabular}
\end{table}

As a baseline, pure MGGP is used (see Section \ref{sec:mggp}).
Our proposed modifications then build upon the baseline by introducing the LCFs\footnote{
    However, the original, unmodified variables are still available to the algorithm.
} and handling them in the way described by the codename.

\paragraph{Example} A configuration with codename UC is a configuration where LCFs are unsynchronized and both backpropagation and mutation are used for tuning the LCF weights.
In one more example, configuration SM is such that LCFs are synchronized and the weights are tuned using only mutation.

\subsection{Parameters} \label{sec:experiments:params}
The new features are, naturally, accompanied by a number of parameters that configure how exactly do those features behave.
Table \ref{tab:parameters} presents all of the parameters, including the ones that come from the baseline algorithm, as well as their values.
The values are based on the default values provided by the GPTIPS2 \cite{gptips-source} package.
The values of parameters related to the LCFs were chosen by hand based on preliminary examination of the configurations.
Except for configurations using backpropagation, we kept the MGGP related parameters for all other configurations too.
For the case of backpropagation, we decided to half the size of population because more time is spent with each individual by tuning the LCF parameters so this enables the algorithm to do more structural exploration.
\begin{table*}[htb]
    \centering
    \caption{
        All parameters of the algorithm and the values we used for them.
        Numbers in parentheses for $|P|$, $|T|$ and $E$ are used for high-level configurations with backpropagation.
        For the purposes of computing $Pr_{Sm}$, $Pr_{Wm}$ is considered zero when no weights mutation is present in the algorithm.
    }
    \label{tab:parameters}
    \begin{tabular}{ccp{123mm}c}
        \toprule
        & parameter & description & value \\\midrule
        \multirow{14}{*}{\vrt{MGGP params}} & $G_{max}$ & maximum number of genes & 10 \\
        & $N_{max}$ & maximum number of nodes per gene & $\infty$ \\
        & $D_{max}$ & maximum depth of a gene & 11 \\
        & $|P|$ & number of individuals in the population & 100 (50)\\
        & $|T|$ & the size of the tournament & 10 (5) \\
        & $E$ & number of top individuals copied to the next generation without any modification (elitism) & 15 (8) \\
        & $Pr_{x}$ & probability of a crossover event & 0.84 \\
        & $Pr_{m}$ & probability of a mutation event & 0.14 \\
        & $Pr_{LLx}$ & probability of a crossover being the low-level one & $1 - Pr_{HLx}$ \\
        & $Pr_{HLx}$ & probability of a crossover being the high-level one & 0.2 \\
        & $r_{HLx}$ & probability that a gene will be selected in high-level crossover & 0.5 \\
        & $Pr_{Sm}$ & probability of a mutation being a subtree mutation & $1 - Pr_{Cm} - Pr_{Wm}$ \\
        & $Pr_{Cm}$ & probability of a mutation being a mutation of constant leaf node(s) & 0.05 \\
        & $\sigma_{Cm}$ & variance of the gaussian distribution used in~constant leaf node mutation & 0.1 \\
        \midrule
        \multirow{4}{*}{\vrt{LCF params}} & $Pr_{Wm}$ & probability of a mutation being a mutation of weights in an LCF & 0.05 \\
        & $\sigma_{Wm}$ & variance of the gaussian distribution used in~weights mutation & 3 \\
        & $Bp_{steps}$ & the number of backprop.-update steps as $Bp_{steps}$ minus the number of nodes in all genes combined & 25 \\
        & $Bp_{min}$ & minimum number of backprop.-update steps each generation per individual (overrides $Bp_{steps}$) & 2 \\
        \bottomrule
    \end{tabular}
\end{table*}

\paragraph{Function set} The functions available to the algorithm were: $+$, $-$, $\times$, sin, cos, exp, $\frac{1}{1 + \mathrm{e}^x}$, tanh, $\frac{\sin x}{x}$, $\ln (1 + \mathrm{e}^x)$, $\mathrm{e}^{-x^2}$ and 2nd to 6th powers.

\subsection{Testing benchmarks and environment} \label{sec:experiments:testing}
We have designed two sets of tests.
First we test the LCF concept on a simple toy problem.
Second we test the overall performance with 9 full benchmarks, both real-world and artificial ones.

\paragraph{Toy problems}
\textbf{S2D}, \textbf{S5D}, \textbf{S10D}, \textbf{RS2D}, \textbf{RS5D}, \textbf{RS10D} are the toy problems in 2, 5 and 10 dimensions.
The S* problems are a simple sigmoid function applied to the first variable, independent on the others.
In the RS* problems, the sigmoid is rotated by $\frac{\pi}{4}$ in all pairs of axes, i.e. all variables are important.
The problems are uniformly randomly sampled from the interval $[-10, 10]^D$.
There are $100\cdot D$ samples in the training set and $250\cdot D$ samples in the testing set.

The goal of these benchmarks is to support our expectation that the LCFs can provide means to find a suitable linear transformation of the feature space.

\paragraph{Realistic problems}
\textbf{K11C} is similar to Keijzer11 in \cite{McDermott2012} but with added numerical coefficients throughout the formula\footnote{
    $f(\mathbf{x}) = (27.22x_1 - 4.54)(-0.39x_2) + 11.46\sin\left(\left(0.21x_1 - 1\right)\left(x_2 + 16.6\right) + 1.97\right)$
}.
The training set is 500 uniform samples from $[-3, 3]^2$, the testing set is a grid in the same range with a spacing of 0.01 in each dimension (361201 samples).

\textbf{UB5D} (Unwrapped Ball 5D) \cite{Vladislavleva2009} is a 5D artificial benchmark.
The true relationship is $f(\mathbf{x}) = \frac{10}{5 + \sum_{i = 1}^{N} \left(x_i - 3\right)^2}$ where $N = 5$.
The training set is 1024 uniform samples from $[-0.25, 6.35]^5$, the testing set is 5000 uniform samples from the same range.

\textbf{ASN} (Airforil Self-Noise), acquired from the UCI repository \cite{uci}, is a 5D dataset regarding the sound pressure levels of airfoils based on measurements from a wind tunnel.
Training/testing set comes from a random 0.7/0.3 split (1503 samples in total).

\textbf{CCS} (Concrete Compressive Strength) \cite{Yeh1998}, acquired from the UCI repository \cite{uci}, is an 8D dataset representing a highly non-linear function of concrete age and ingredients.
Training/testing set comes from a random 0.7/0.3 split (1030 samples in total).

\textbf{ENC} and \textbf{ENH} (Energy Efficiency) \cite{Tsanas2012}, acquired from the UCI repository \cite{uci}, are 8D datasets regarding the energy efficiency of cooling and heating of buildings.
Training/testing set comes from a random 0.7/0.3 split (768 samples in total per dataset).

\textbf{SU}\footnote{
    \label{footnote:rl-datasets}
    These datasets are depicted in the supplementary material.
} and \textbf{SU-I}\footnotemark[7] (swingup) are two datasets from a reinforcement learning domain.
They are value functions of an inverted pendulum swing-up problem computed by a numeric approximator.
They have 2 dimensions (pendulum angle and angular velocity) and the value is the value of the state w.r.t. the goal state which, for the SU variant, is located at $[-\pi, 0]$ and equivalently $[\pi, 0]$ (due to the circular nature of the problem).
The SU-I variant represents identical function but the angle coordinate is shifted by $\frac{\pi}{2}$.
Training/testing set comes from a random 0.7/0.3 split (441 samples in total).

\textbf{MM}\footnotemark[7] (2-coil magnetic manipulation) is a dataset from a reinforcement learning domain.
It is a value function of a linear magnetic manipulation problem with 2 coils computed by a numeric approximator.
It has 2 dimensions (the manipulated ball's position and velocity) and the value is the value of the state w.r.t. the goal state.
Training/testing set comes from a random 0.7/0.3 split (729 samples in total).

\paragraph{Testing methodology} For each benchmark, each algorithm configuration was run 30 times, each time with a different seed and different sampling (for the artificial benchmarks) or different training/testing split (for the real-world benchmarks).
Each run has a wall-clock time limit of 7 minutes and is terminated after this amount of time passes (except up to a negligible amount of time that passes between the time checks).
Fitness is $R^2$ on the training dataset and is maximized.
After the algorithm finishes, the resulting model (the one with the best fitness found over the whole runtime of the algorithm) is evaluated on the testing set.
In the next section we report the results both on the training and testing sets so that some judgement on overfitting can be made.

\paragraph{Testing environment} Everything is written in Python 3 and uses the NumPy library for vector and matrix calculations including the linear regression.
All experiments were carried out on the National Grid Infrastructure MetaCentrum (see Acknowledgements) which is a heterogeneous computational grid.
We ensured that for each dataset all runs of all configurations on that dataset were carried out on machines of the same cluster which all have the same configuration.

\section{Results and discussion} \label{sec:results}
In this section we present the results of experiments we described in the previous section.

\subsection{Toy problems}
Here we present the results on the toy problems S5D and RS5D.
The results can be seen in Tables \ref{tab:s5d} and \ref{tab:rs5d}.
Results on (R)S2D and (R)S10D are not presented because 2D case was too easy for all algorithms, and the results on 10D case are very similar to the presented 5D case.

The column denoted ``vb'', which stands for ``versus baseline'', shows whether the configuration was better than (denoted by \cmark), worse than (denoted by \xmark) or indifferent to (denoted by blank space) the baseline.
This result was established using Mann-Whitney ranksum test on the testing $R^2$ values with the significance level $\alpha = 0.05$ with Bonferonni correction for 11 comparisons\footnote{
    We present only 8 here because the randomized initialization (see footnote~\ref{footnote:random}) we dismissed because of no impact but three such configurations were part of this comparison.
} resulting in an effective $\alpha \approx 0.0045$.

The column denoted ``LCF'' shows mean fraction of non-constant leaf nodes that are LCFs\footnote{
    For example, expression containing 3 ``pure'' variables and 7 LCFs would have this value equal to 0.7.
}.
\begin{table}[h!tb]
    \centering
    \caption{Results on the S5D toy problem.}
    \label{tab:s5d}
    \begin{tabular}{c@{\hspace{0mm}}cccccccc}
        \toprule
        \multirow{2}{*}[-2.5mm]{\vrt{\scriptsize mode}} & \multirow{2}{*}[-2mm]{\vrt{\scriptsize tuning}} & \multicolumn{2}{c}{training R$^2$} & \multicolumn{2}{c}{testing R$^2$} & & mean & mean \\
        \cmidrule(lr){3-4} \cmidrule(lr){5-6}
        & & median & $\begin{smallmatrix}\text{\scriptsize max}\\\text{\scriptsize min}\end{smallmatrix}$ & median & $\begin{smallmatrix}\text{\scriptsize max}\\\text{\scriptsize min}\end{smallmatrix}$ & vb & LCF & depth \\
        \midrule
        - & - & 1 & $\begin{smallmatrix}{1}\\{1}\end{smallmatrix}$ & 1 & $\begin{smallmatrix}{1}\\{1}\end{smallmatrix}$ &   & 0 & 4.33 \\[1.5mm]
        U & M & 1 & $\begin{smallmatrix}{1}\\{1}\end{smallmatrix}$ & 1 & $\begin{smallmatrix}{1}\\{1}\end{smallmatrix}$ &   & 0.482 & 4.1 \\[1.5mm]
        U & B & 1 & $\begin{smallmatrix}{1}\\{1}\end{smallmatrix}$ & 1 & $\begin{smallmatrix}{1}\\{1}\end{smallmatrix}$ &   & 0.576 & 3.5 \\[1.5mm]
        U & C & 1 & $\begin{smallmatrix}{1}\\{1}\end{smallmatrix}$ & 1 & $\begin{smallmatrix}{1}\\{1}\end{smallmatrix}$ &   & 0.521 & 3.77 \\[1.5mm]
        S & M & 1 & $\begin{smallmatrix}{1}\\{1}\end{smallmatrix}$ & 1 & $\begin{smallmatrix}{1}\\{1}\end{smallmatrix}$ &   & 0.469 & 4.13 \\[1.5mm]
        S & B & 1 & $\begin{smallmatrix}{1}\\{1}\end{smallmatrix}$ & 1 & $\begin{smallmatrix}{1}\\{1}\end{smallmatrix}$ &   & 0.475 & 4.2 \\[1.5mm]
        S & C & 1 & $\begin{smallmatrix}{1}\\{1}\end{smallmatrix}$ & 1 & $\begin{smallmatrix}{1}\\{1}\end{smallmatrix}$ &   & 0.48 & 3.73 \\[1.5mm]
        G & B & 1 & $\begin{smallmatrix}{1}\\{1}\end{smallmatrix}$ & 1 & $\begin{smallmatrix}{1}\\{1}\end{smallmatrix}$ &   & 0.38 & 4.23 \\[1.5mm]
        G & C & 1 & $\begin{smallmatrix}{1}\\{1}\end{smallmatrix}$ & 1 & $\begin{smallmatrix}{1}\\{1}\end{smallmatrix}$ &   & 0.352 & 4.87 \\[1.5mm]
        \bottomrule
    \end{tabular}
\end{table}
\begin{table}[h!tb]
    \centering
    \caption{Results on the RS5D toy problem.}
    \label{tab:rs5d}
    \begin{tabular}{c@{\hspace{0mm}}cccccccc}
        \toprule
        \multirow{2}{*}[-2.5mm]{\vrt{\scriptsize mode}} & \multirow{2}{*}[-2mm]{\vrt{\scriptsize tuning}} & \multicolumn{2}{c}{training R$^2$} & \multicolumn{2}{c}{testing R$^2$} & & mean & mean \\
        \cmidrule(lr){3-4} \cmidrule(lr){5-6}
        & & median & $\begin{smallmatrix}\text{\scriptsize max}\\\text{\scriptsize min}\end{smallmatrix}$ & median & $\begin{smallmatrix}\text{\scriptsize max}\\\text{\scriptsize min}\end{smallmatrix}$ & vb & LCF & depth \\
        \midrule
        - & - & 0.995 & $\begin{smallmatrix}{0.997}\\{0.942}\end{smallmatrix}$ & 0.991 & $\begin{smallmatrix}{0.996}\\{0.912}\end{smallmatrix}$ &   & 0 & 10.9 \\[1.5mm]
        U & M & 0.993 & $\begin{smallmatrix}{1}\\{0.93}\end{smallmatrix}$ & 0.99 & $\begin{smallmatrix}{1}\\{0.907}\end{smallmatrix}$ &   & 0.524 & 10.6 \\[1.5mm]
        U & B & 1 & $\begin{smallmatrix}{1}\\{1}\end{smallmatrix}$ & 1 & $\begin{smallmatrix}{1}\\{1}\end{smallmatrix}$ & \cmark & 0.98 & 4.6 \\[1.5mm]
        U & C & 1 & $\begin{smallmatrix}{1}\\{1}\end{smallmatrix}$ & 1 & $\begin{smallmatrix}{1}\\{1}\end{smallmatrix}$ & \cmark & 0.974 & 4.63 \\[1.5mm]
        S & M & 0.995 & $\begin{smallmatrix}{1}\\{0.158}\end{smallmatrix}$ & 0.993 & $\begin{smallmatrix}{0.999}\\{-0.168}\end{smallmatrix}$ &   & 0.579 & 10.9 \\[1.5mm]
        S & B & 1 & $\begin{smallmatrix}{1}\\{1}\end{smallmatrix}$ & 1 & $\begin{smallmatrix}{1}\\{1}\end{smallmatrix}$ & \cmark & 0.9 & 8.1 \\[1.5mm]
        S & C & 1 & $\begin{smallmatrix}{1}\\{1}\end{smallmatrix}$ & 1 & $\begin{smallmatrix}{1}\\{1}\end{smallmatrix}$ & \cmark & 0.954 & 7.77 \\[1.5mm]
        G & B & 1 & $\begin{smallmatrix}{1}\\{0.855}\end{smallmatrix}$ & 1 & $\begin{smallmatrix}{1}\\{0.659}\end{smallmatrix}$ & \cmark & 0.817 & 8.53 \\[1.5mm]
        G & C & 0.974 & $\begin{smallmatrix}{1}\\{0.872}\end{smallmatrix}$ & 0.962 & $\begin{smallmatrix}{1}\\{0.736}\end{smallmatrix}$ & \xmark & 0.651 & 6.7 \\[1.5mm]
        \bottomrule
    \end{tabular}
\end{table}

\paragraph{Discussion}
The toy problems showed that the basic idea is supported -- unrotated problems were easy for all configurations while the rotated ones were easy for the configurations using LCFs and not for baseline.
We can see that the configurations that were statistically better than the baseline show high LCF ratio, i.e. they really use the LCFs.

However, we can see that UM and SM, i.e. configurations with LCF weights tuned only by mutation, were (statistically) neither better, nor worse than the baseline (though the maximum achieved values are better).
We can also see that these configurations have much smaller LCF ratio than the other LCF-enabled configurations.
This can be explained as the lack of the ability to tune the LCF weights accurately enough to be beneficial for the expressions.
This can be seen as an indication that mutation is not very good approach to search for the linear combinations.

Note that the LCFs are also used in the non-rotated problem although they are of no benefit there.
This is due to two facts: (i) there is no penalty for using them, so they can be used similarly to ordinary variables, and (ii) the genes containing the LCFs can receive a negligible coefficient in the top-level linear combination in case the true relationship is among the genes (that one will receive a coeffcient of 1) so they could be considered as effectively not present.

Also interesting is the depth usage.
On the rotated problem, the baseline uses almost all the available depth while UB, for example, uses less than half the depth\footnote{
    Additionally, similar effect as in the previous paragraph can apply -- some deeper genes can have a small top-level linear coefficient than other gene(s).
}.
This is a clear indication of the ability of LCFs to provide good data to the rest of the expressions.

\subsection{Realistic problems}
First we present a summary result of which configurations were better than the baseline across individual datasets.
The result can be seen in Table \ref{tab:summary}.
The table shows the number of datasets where each configuration was better than, tied with, or worse than the baseline.
This comparison was established using identical test as the ``versus baseline'' test in previous subsection.
\begin{table}[htb]
    \centering
    \caption{
        Summary results for each algorithm configuration.
    }
    \label{tab:summary}
    \begin{tabular}{rcccccccc}
        \toprule
        mode         & U & U & U & S & S & S & G & G \\[-1mm]
        tuning       & M & B & C & M & B & C & B & C \\
        \midrule
        better than baseline & 1 & 5 & 5 & 1 & 3 & 3 &   &   \\
        indifferent to baseline & 8 & 3 & 3 & 8 & 4 & 4 & 6 & 3 \\
        worse than baseline &   & 1 & 1 &   & 2 & 2 & 3 & 6 \\
        \bottomrule
    \end{tabular}
\end{table}

\begin{figure}[htb]
    \includegraphics[width=0.37\textwidth,trim=0mm 4.4mm 2mm 2.5mm,clip]{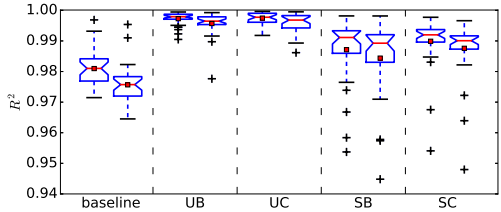}
    \caption{
        Training (left) and testing (right) $R^2$ performance on the K11C dataset.
        Note that for the testing UC plot, one outlier at \mbox{-3.24e29} is not shown (as well as the mean marker).
    }
    \label{fig:k11c}
\end{figure}

\begin{figure}[htb]
    \includegraphics[width=0.37\textwidth,trim=0mm 4.4mm 2mm 2.5mm,clip]{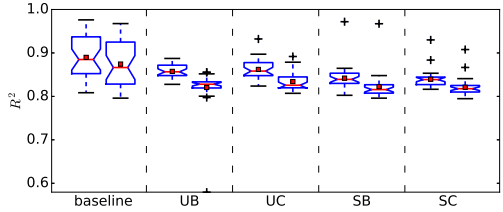}
    \caption{
        Training (left) and testing (right) $R^2$ performance on the UB5D dataset.
    }
    \label{fig:ub5d}
\end{figure}

\begin{figure}[htb]
    \includegraphics[width=0.37\textwidth,trim=0mm 4.4mm 2mm 2.5mm,clip]{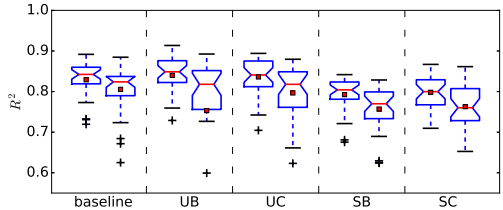}
    \caption{
        Training (left) and testing (right) $R^2$ performance on the ASN dataset.
        Note that for the testing UB plot, one outlier at -0.719 is not shown.
    }
    \label{fig:asn}
\end{figure}

\begin{figure}[htb]
    \includegraphics[width=0.37\textwidth,trim=0mm 4.4mm 2mm 2.5mm,clip]{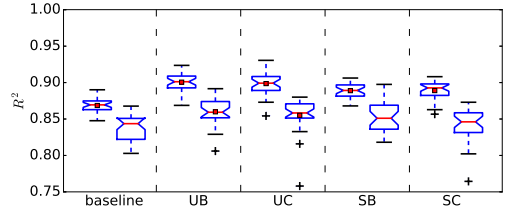}
    \caption{
        Training (left) and testing (right) $R^2$ performance on the CCS dataset.
        Note that for baseline, SB and SC, testing outliers (one configuration each) at \mbox{-8.68e7}, \mbox{-4.74e4} and \mbox{-291} respectively are not shown.
    }
    \label{fig:ccs}
\end{figure}

\begin{figure}[htb]
    \includegraphics[width=0.37\textwidth,trim=0mm 4.4mm 2mm 2.5mm,clip]{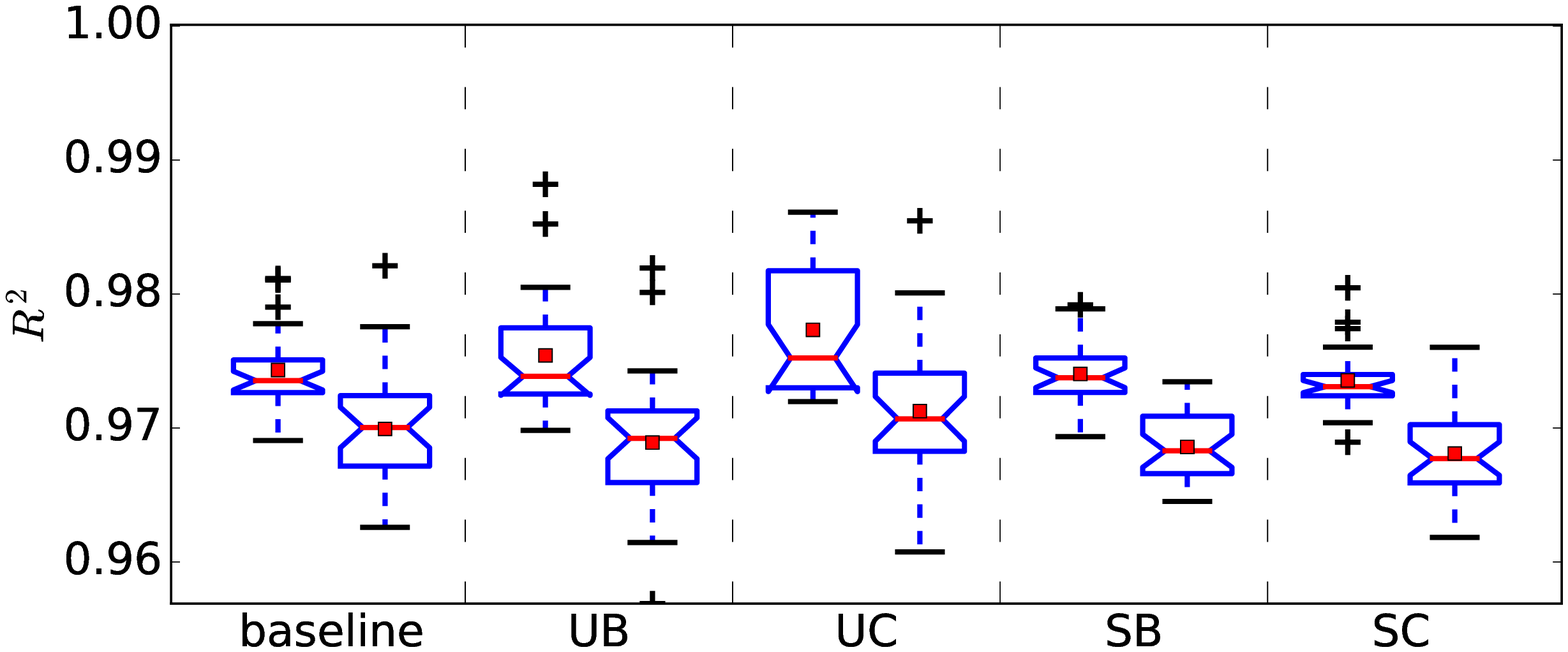}
    \caption{
        Training (left) and testing (right) $R^2$ performance on the ENC dataset.
    }
    \label{fig:enc}
\end{figure}

\begin{figure}[htb]
    \includegraphics[width=0.37\textwidth,trim=0mm 4.4mm 2mm 2.5mm,clip]{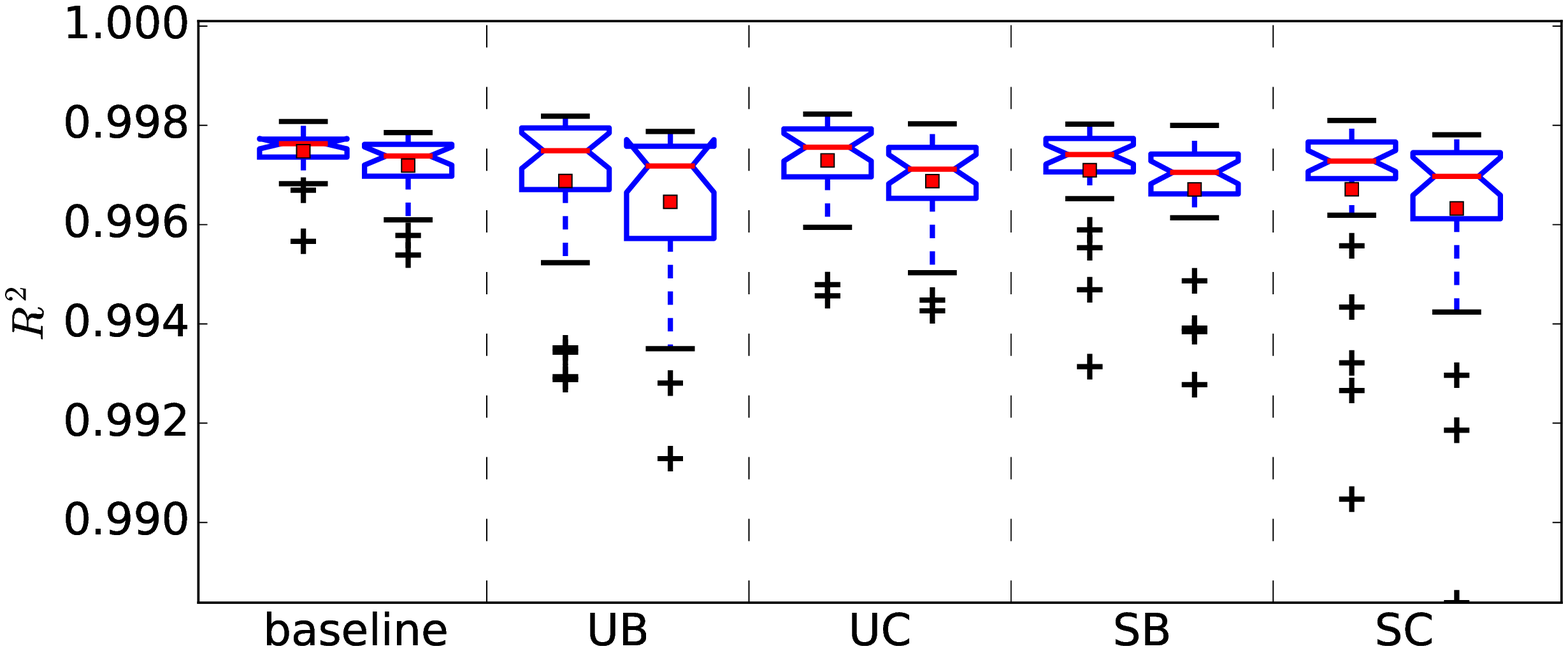}
    \caption{
        Training (left) and testing (right) $R^2$ performance on the ENH dataset.
    }
    \label{fig:enh}
\end{figure}

\begin{figure}[htb]
    \includegraphics[width=0.37\textwidth,trim=0mm 4.4mm 2mm 2.5mm,clip]{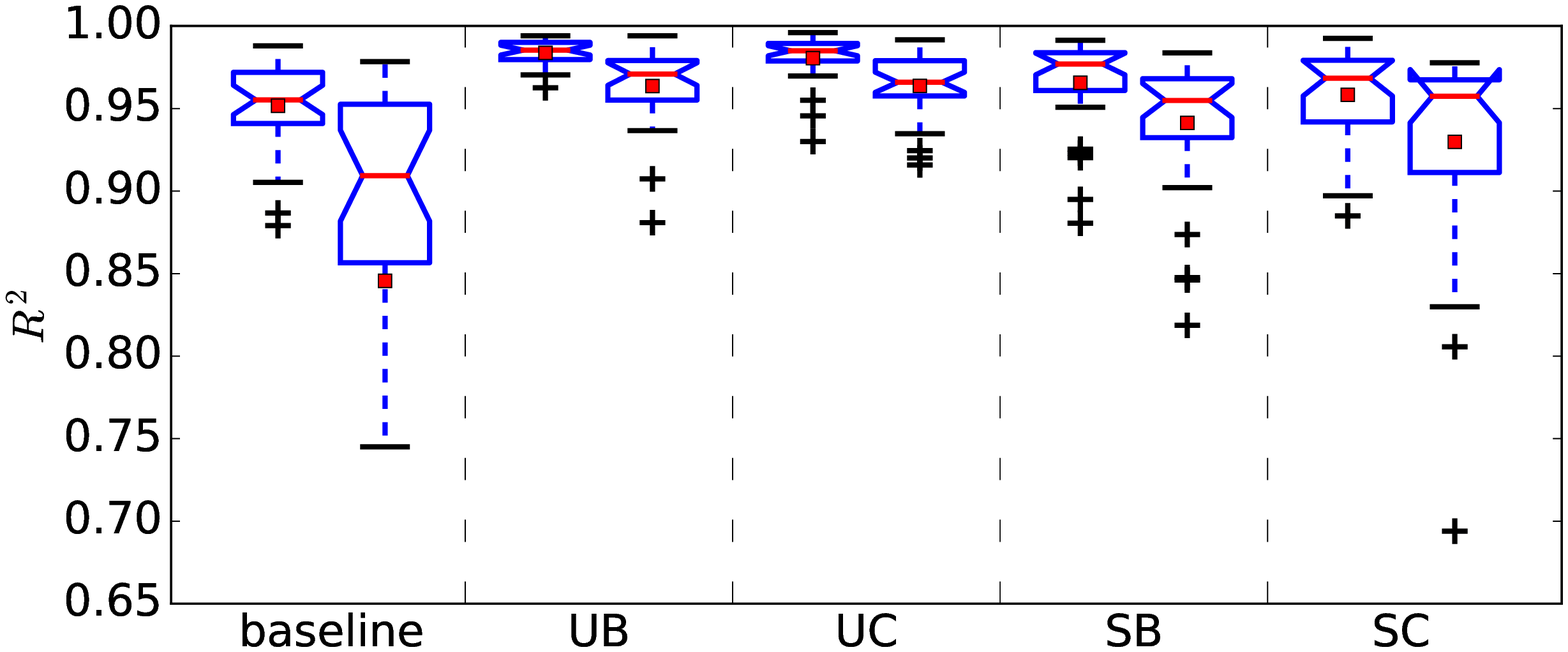}
    \caption{
        Training (left) and testing (right) $R^2$ performance on the SU dataset.
        Note that for baseline, one testing outlier at \mbox{-0.664} is not shown.
    }
    \label{fig:su}
\end{figure}

\begin{figure}[htb]
    \includegraphics[width=0.37\textwidth,trim=0mm 4.4mm 2mm 2.5mm,clip]{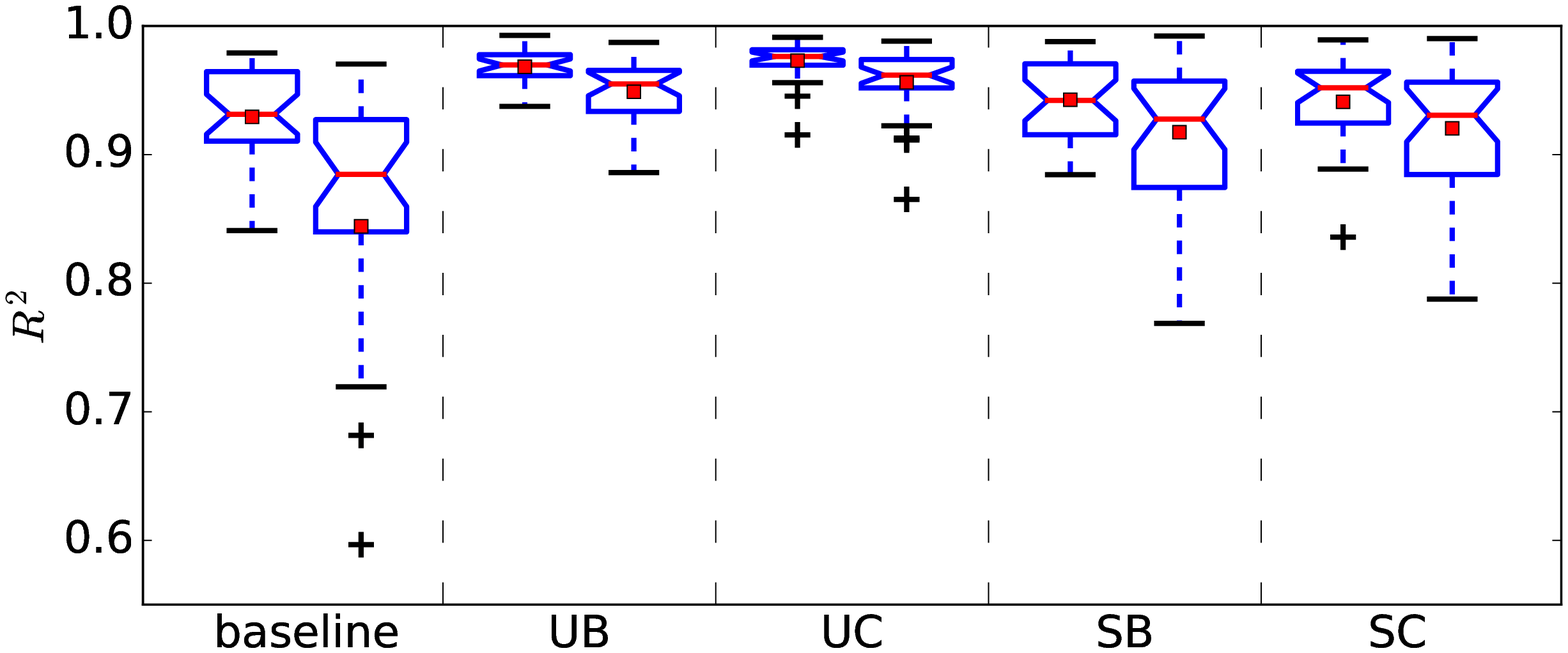}
    \caption{
        Training (left) and testing (right) $R^2$ performance on the SU-I dataset.
        Note that for baseline, one testing outlier at \mbox{0.175} is not shown.
    }
    \label{fig:su-i}
\end{figure}

\begin{figure}[htb]
    \includegraphics[width=0.37\textwidth,trim=0mm 4.4mm 2mm 2.5mm,clip]{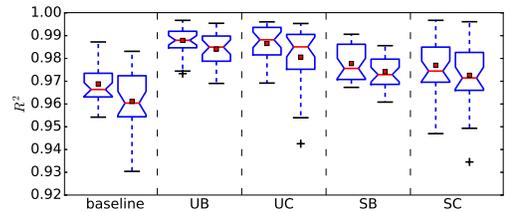}
    \caption{
        Training (left) and testing (right) $R^2$ performance on the MM dataset.
        Note that for baseline, one testing outlier at \mbox{0.175} is not shown.
    }
    \label{fig:mm}
\end{figure}

We will now present more detailed results of the configurations for each dataset.
We will list only results for configurations UB, UC, SB and SC as these had the most positive results in Table \ref{tab:summary}.
Full results are available in the supplementary material to this article.
The results will be presented in the form of tables (see Tables \ref{tab:k11c} through \ref{tab:mm}) and box plots (see Figures \ref{fig:k11c} through \ref{fig:mm}).
The tables are identical to tables for toy problems, with exactly the same meaning of columns, except we don't present the mean depth here.

\paragraph{Negative $R^2$}
We can see that in some cases, some configurations achieved a negative $R^2$ on the testing set\footnote{
    Negative $R^2$ means that the fit is worse than that of the constant model equal to the mean of the target data.
}.
In all of those cases, it happened in only a single run of the 30 runs for that configuration and dataset.

\begin{table}[ht]
    \centering
    \caption{Performance on the K11C dataset.}
    \label{tab:k11c}
    \begin{tabular}{c@{\hspace{0mm}}ccccccc}
        \toprule
        \multirow{2}{*}[-2.5mm]{\vrt{\scriptsize mode}} & \multirow{2}{*}[-2mm]{\vrt{\scriptsize tuning}} & \multicolumn{2}{c}{training R$^2$} & \multicolumn{2}{c}{testing R$^2$} & & mean \\
        \cmidrule(lr){3-4} \cmidrule(lr){5-6}
        & & median & $\begin{smallmatrix}\text{\scriptsize max}\\\text{\scriptsize min}\end{smallmatrix}$ & median & $\begin{smallmatrix}\text{\scriptsize max}\\\text{\scriptsize min}\end{smallmatrix}$ & vb & LCF \\
        \midrule
        - & - & 0.981 & $\begin{smallmatrix}{0.997}\\{0.971}\end{smallmatrix}$ & 0.976 & $\begin{smallmatrix}{0.995}\\{0.965}\end{smallmatrix}$ &   & 0 \\[1.5mm]
        U & B & 0.998 & $\begin{smallmatrix}{0.999}\\{0.99}\end{smallmatrix}$ & 0.996 & $\begin{smallmatrix}{0.999}\\{0.978}\end{smallmatrix}$ & \cmark & 0.873 \\[1.5mm]
        U & C & 0.998 & $\begin{smallmatrix}{1}\\{0.992}\end{smallmatrix}$ & 0.997 & $\begin{smallmatrix}{1}\\{-3.24e+29}\end{smallmatrix}$ & \cmark & 0.874 \\[1.5mm]
        S & B & 0.991 & $\begin{smallmatrix}{0.998}\\{0.954}\end{smallmatrix}$ & 0.989 & $\begin{smallmatrix}{0.998}\\{0.945}\end{smallmatrix}$ & \cmark & 0.603 \\[1.5mm]
        S & C & 0.992 & $\begin{smallmatrix}{0.998}\\{0.954}\end{smallmatrix}$ & 0.99 & $\begin{smallmatrix}{0.997}\\{0.948}\end{smallmatrix}$ & \cmark & 0.622 \\[1.5mm]
        \bottomrule
    \end{tabular}
\end{table}
\begin{table}[ht]
    \centering
    \caption{Performance on the UB5D dataset.}
    \label{tab:ub5d}
    \begin{tabular}{c@{\hspace{0mm}}ccccccc}
        \toprule
        \multirow{2}{*}[-2.5mm]{\vrt{\scriptsize mode}} & \multirow{2}{*}[-2mm]{\vrt{\scriptsize tuning}} & \multicolumn{2}{c}{training R$^2$} & \multicolumn{2}{c}{testing R$^2$} & & mean \\
        \cmidrule(lr){3-4} \cmidrule(lr){5-6}
        & & median & $\begin{smallmatrix}\text{\scriptsize max}\\\text{\scriptsize min}\end{smallmatrix}$ & median & $\begin{smallmatrix}\text{\scriptsize max}\\\text{\scriptsize min}\end{smallmatrix}$ & vb & LCF \\
        \midrule
        - & - & 0.885 & $\begin{smallmatrix}{0.976}\\{0.808}\end{smallmatrix}$ & 0.866 & $\begin{smallmatrix}{0.968}\\{0.796}\end{smallmatrix}$ &   & 0 \\[1.5mm]
        U & B & 0.857 & $\begin{smallmatrix}{0.887}\\{0.828}\end{smallmatrix}$ & 0.828 & $\begin{smallmatrix}{0.856}\\{0.58}\end{smallmatrix}$ & \xmark & 0.823 \\[1.5mm]
        U & C & 0.858 & $\begin{smallmatrix}{0.932}\\{0.824}\end{smallmatrix}$ & 0.826 & $\begin{smallmatrix}{0.892}\\{0.807}\end{smallmatrix}$ & \xmark & 0.802 \\[1.5mm]
        S & B & 0.839 & $\begin{smallmatrix}{0.972}\\{0.802}\end{smallmatrix}$ & 0.816 & $\begin{smallmatrix}{0.967}\\{0.796}\end{smallmatrix}$ & \xmark & 0.553 \\[1.5mm]
        S & C & 0.839 & $\begin{smallmatrix}{0.93}\\{0.816}\end{smallmatrix}$ & 0.818 & $\begin{smallmatrix}{0.908}\\{0.795}\end{smallmatrix}$ & \xmark & 0.601 \\[1.5mm]
        \bottomrule
    \end{tabular}
\end{table}
\begin{table}[ht]
    \centering
    \caption{Performance on the ASN dataset.}
    \label{tab:asn}
    \begin{tabular}{c@{\hspace{0mm}}ccccccc}
        \toprule
        \multirow{2}{*}[-2.5mm]{\vrt{\scriptsize mode}} & \multirow{2}{*}[-2mm]{\vrt{\scriptsize tuning}} & \multicolumn{2}{c}{training R$^2$} & \multicolumn{2}{c}{testing R$^2$} & & mean \\
        \cmidrule(lr){3-4} \cmidrule(lr){5-6}
        & & median & $\begin{smallmatrix}\text{\scriptsize max}\\\text{\scriptsize min}\end{smallmatrix}$ & median & $\begin{smallmatrix}\text{\scriptsize max}\\\text{\scriptsize min}\end{smallmatrix}$ & vb & LCF \\
        \midrule
        - & - & 0.842 & $\begin{smallmatrix}{0.892}\\{0.72}\end{smallmatrix}$ & 0.824 & $\begin{smallmatrix}{0.885}\\{0.625}\end{smallmatrix}$ &   & 0 \\[1.5mm]
        U & B & 0.849 & $\begin{smallmatrix}{0.914}\\{0.729}\end{smallmatrix}$ & 0.818 & $\begin{smallmatrix}{0.893}\\{-0.719}\end{smallmatrix}$ &   & 0.834 \\[1.5mm]
        U & C & 0.841 & $\begin{smallmatrix}{0.894}\\{0.705}\end{smallmatrix}$ & 0.818 & $\begin{smallmatrix}{0.88}\\{0.623}\end{smallmatrix}$ &   & 0.828 \\[1.5mm]
        S & B & 0.804 & $\begin{smallmatrix}{0.842}\\{0.675}\end{smallmatrix}$ & 0.77 & $\begin{smallmatrix}{0.829}\\{0.624}\end{smallmatrix}$ & \xmark & 0.651 \\[1.5mm]
        S & C & 0.8 & $\begin{smallmatrix}{0.867}\\{0.71}\end{smallmatrix}$ & 0.76 & $\begin{smallmatrix}{0.861}\\{0.653}\end{smallmatrix}$ & \xmark & 0.68 \\[1.5mm]
        \bottomrule
    \end{tabular}
\end{table}
\begin{table}[ht]
    \centering
    \caption{Performance on the CCS dataset.}
    \label{tab:ccs}
    \begin{tabular}{c@{\hspace{0mm}}ccccccc}
        \toprule
        \multirow{2}{*}[-2.5mm]{\vrt{\scriptsize mode}} & \multirow{2}{*}[-2mm]{\vrt{\scriptsize tuning}} & \multicolumn{2}{c}{training R$^2$} & \multicolumn{2}{c}{testing R$^2$} & & mean \\
        \cmidrule(lr){3-4} \cmidrule(lr){5-6}
        & & median & $\begin{smallmatrix}\text{\scriptsize max}\\\text{\scriptsize min}\end{smallmatrix}$ & median & $\begin{smallmatrix}\text{\scriptsize max}\\\text{\scriptsize min}\end{smallmatrix}$ & vb & LCF \\
        \midrule
        - & - & 0.869 & $\begin{smallmatrix}{0.89}\\{0.848}\end{smallmatrix}$ & 0.844 & $\begin{smallmatrix}{0.868}\\{-8.68e+07}\end{smallmatrix}$ &   & 0 \\[1.5mm]
        U & B & 0.901 & $\begin{smallmatrix}{0.924}\\{0.869}\end{smallmatrix}$ & 0.859 & $\begin{smallmatrix}{0.892}\\{0.806}\end{smallmatrix}$ & \cmark & 0.87 \\[1.5mm]
        U & C & 0.899 & $\begin{smallmatrix}{0.931}\\{0.854}\end{smallmatrix}$ & 0.858 & $\begin{smallmatrix}{0.88}\\{0.758}\end{smallmatrix}$ & \cmark & 0.885 \\[1.5mm]
        S & B & 0.889 & $\begin{smallmatrix}{0.906}\\{0.868}\end{smallmatrix}$ & 0.851 & $\begin{smallmatrix}{0.898}\\{-4.74e+04}\end{smallmatrix}$ &   & 0.676 \\[1.5mm]
        S & C & 0.893 & $\begin{smallmatrix}{0.908}\\{0.857}\end{smallmatrix}$ & 0.846 & $\begin{smallmatrix}{0.873}\\{-291}\end{smallmatrix}$ &   & 0.707 \\[1.5mm]
        \bottomrule
    \end{tabular}
\end{table}
\begin{table}[ht]
    \centering
    \caption{Performance on the ENC dataset.}
    \label{tab:enc}
    \begin{tabular}{c@{\hspace{0mm}}ccccccc}
        \toprule
        \multirow{2}{*}[-2.5mm]{\vrt{\scriptsize mode}} & \multirow{2}{*}[-2mm]{\vrt{\scriptsize tuning}} & \multicolumn{2}{c}{training R$^2$} & \multicolumn{2}{c}{testing R$^2$} & & mean \\
        \cmidrule(lr){3-4} \cmidrule(lr){5-6}
        & & median & $\begin{smallmatrix}\text{\scriptsize max}\\\text{\scriptsize min}\end{smallmatrix}$ & median & $\begin{smallmatrix}\text{\scriptsize max}\\\text{\scriptsize min}\end{smallmatrix}$ & vb & LCF \\
        \midrule
        - & - & 0.974 & $\begin{smallmatrix}{0.981}\\{0.969}\end{smallmatrix}$ & 0.97 & $\begin{smallmatrix}{0.982}\\{0.963}\end{smallmatrix}$ &   & 0 \\[1.5mm]
        U & B & 0.974 & $\begin{smallmatrix}{0.988}\\{0.97}\end{smallmatrix}$ & 0.969 & $\begin{smallmatrix}{0.982}\\{0.957}\end{smallmatrix}$ &   & 0.751 \\[1.5mm]
        U & C & 0.975 & $\begin{smallmatrix}{0.986}\\{0.972}\end{smallmatrix}$ & 0.971 & $\begin{smallmatrix}{0.985}\\{0.961}\end{smallmatrix}$ &   & 0.772 \\[1.5mm]
        S & B & 0.974 & $\begin{smallmatrix}{0.979}\\{0.969}\end{smallmatrix}$ & 0.968 & $\begin{smallmatrix}{0.973}\\{0.965}\end{smallmatrix}$ &   & 0.609 \\[1.5mm]
        S & C & 0.973 & $\begin{smallmatrix}{0.98}\\{0.969}\end{smallmatrix}$ & 0.968 & $\begin{smallmatrix}{0.976}\\{0.962}\end{smallmatrix}$ &   & 0.609 \\[1.5mm]
        \bottomrule
    \end{tabular}
\end{table}
\begin{table}[ht]
    \centering
    \caption{Performance on the ENH dataset.}
    \label{tab:enh}
    \begin{tabular}{c@{\hspace{0mm}}ccccccc}
        \toprule
        \multirow{2}{*}[-2.5mm]{\vrt{\scriptsize mode}} & \multirow{2}{*}[-2mm]{\vrt{\scriptsize tuning}} & \multicolumn{2}{c}{training R$^2$} & \multicolumn{2}{c}{testing R$^2$} & & mean \\
        \cmidrule(lr){3-4} \cmidrule(lr){5-6}
        & & median & $\begin{smallmatrix}\text{\scriptsize max}\\\text{\scriptsize min}\end{smallmatrix}$ & median & $\begin{smallmatrix}\text{\scriptsize max}\\\text{\scriptsize min}\end{smallmatrix}$ & vb & LCF \\
        \midrule
        - & - & 0.998 & $\begin{smallmatrix}{0.998}\\{0.996}\end{smallmatrix}$ & 0.997 & $\begin{smallmatrix}{0.998}\\{0.995}\end{smallmatrix}$ &   & 0 \\[1.5mm]
        U & B & 0.997 & $\begin{smallmatrix}{0.998}\\{0.993}\end{smallmatrix}$ & 0.997 & $\begin{smallmatrix}{0.998}\\{0.991}\end{smallmatrix}$ &   & 0.73 \\[1.5mm]
        U & C & 0.998 & $\begin{smallmatrix}{0.998}\\{0.995}\end{smallmatrix}$ & 0.997 & $\begin{smallmatrix}{0.998}\\{0.994}\end{smallmatrix}$ &   & 0.732 \\[1.5mm]
        S & B & 0.997 & $\begin{smallmatrix}{0.998}\\{0.993}\end{smallmatrix}$ & 0.997 & $\begin{smallmatrix}{0.998}\\{0.993}\end{smallmatrix}$ &   & 0.592 \\[1.5mm]
        S & C & 0.997 & $\begin{smallmatrix}{0.998}\\{0.99}\end{smallmatrix}$ & 0.997 & $\begin{smallmatrix}{0.998}\\{0.988}\end{smallmatrix}$ &   & 0.61 \\[1.5mm]
        \bottomrule
    \end{tabular}
\end{table}
\begin{table}[ht]
    \centering
    \caption{Performance on the SU dataset.}
    \label{tab:su}
    \begin{tabular}{c@{\hspace{0mm}}ccccccc}
        \toprule
        \multirow{2}{*}[-2.5mm]{\vrt{\scriptsize mode}} & \multirow{2}{*}[-2mm]{\vrt{\scriptsize tuning}} & \multicolumn{2}{c}{training R$^2$} & \multicolumn{2}{c}{testing R$^2$} & & mean \\
        \cmidrule(lr){3-4} \cmidrule(lr){5-6}
        & & median & $\begin{smallmatrix}\text{\scriptsize max}\\\text{\scriptsize min}\end{smallmatrix}$ & median & $\begin{smallmatrix}\text{\scriptsize max}\\\text{\scriptsize min}\end{smallmatrix}$ & vb & LCF \\
        \midrule
        - & - & 0.955 & $\begin{smallmatrix}{0.988}\\{0.879}\end{smallmatrix}$ & 0.909 & $\begin{smallmatrix}{0.978}\\{-0.664}\end{smallmatrix}$ &   & 0 \\[1.5mm]
        U & B & 0.985 & $\begin{smallmatrix}{0.994}\\{0.963}\end{smallmatrix}$ & 0.971 & $\begin{smallmatrix}{0.994}\\{0.881}\end{smallmatrix}$ & \cmark & 0.894 \\[1.5mm]
        U & C & 0.985 & $\begin{smallmatrix}{0.996}\\{0.93}\end{smallmatrix}$ & 0.966 & $\begin{smallmatrix}{0.992}\\{0.916}\end{smallmatrix}$ & \cmark & 0.885 \\[1.5mm]
        S & B & 0.977 & $\begin{smallmatrix}{0.991}\\{0.881}\end{smallmatrix}$ & 0.955 & $\begin{smallmatrix}{0.984}\\{0.819}\end{smallmatrix}$ & \cmark & 0.598 \\[1.5mm]
        S & C & 0.968 & $\begin{smallmatrix}{0.993}\\{0.885}\end{smallmatrix}$ & 0.958 & $\begin{smallmatrix}{0.978}\\{0.694}\end{smallmatrix}$ &   & 0.633 \\[1.5mm]
        \bottomrule
    \end{tabular}
\end{table}
\begin{table}[ht]
    \centering
    \caption{Performance on the SU-I dataset.}
    \label{tab:su-i}
    \begin{tabular}{c@{\hspace{0mm}}ccccccc}
        \toprule
        \multirow{2}{*}[-2.5mm]{\vrt{\scriptsize mode}} & \multirow{2}{*}[-2mm]{\vrt{\scriptsize tuning}} & \multicolumn{2}{c}{training R$^2$} & \multicolumn{2}{c}{testing R$^2$} & & mean \\
        \cmidrule(lr){3-4} \cmidrule(lr){5-6}
        & & median & $\begin{smallmatrix}\text{\scriptsize max}\\\text{\scriptsize min}\end{smallmatrix}$ & median & $\begin{smallmatrix}\text{\scriptsize max}\\\text{\scriptsize min}\end{smallmatrix}$ & vb & LCF \\
        \midrule
        - & - & 0.931 & $\begin{smallmatrix}{0.979}\\{0.841}\end{smallmatrix}$ & 0.885 & $\begin{smallmatrix}{0.97}\\{0.175}\end{smallmatrix}$ &   & 0 \\[1.5mm]
        U & B & 0.97 & $\begin{smallmatrix}{0.993}\\{0.938}\end{smallmatrix}$ & 0.955 & $\begin{smallmatrix}{0.987}\\{0.886}\end{smallmatrix}$ & \cmark & 0.895 \\[1.5mm]
        U & C & 0.976 & $\begin{smallmatrix}{0.991}\\{0.915}\end{smallmatrix}$ & 0.962 & $\begin{smallmatrix}{0.988}\\{0.865}\end{smallmatrix}$ & \cmark & 0.912 \\[1.5mm]
        S & B & 0.942 & $\begin{smallmatrix}{0.988}\\{0.884}\end{smallmatrix}$ & 0.928 & $\begin{smallmatrix}{0.992}\\{0.769}\end{smallmatrix}$ &   & 0.569 \\[1.5mm]
        S & C & 0.952 & $\begin{smallmatrix}{0.989}\\{0.836}\end{smallmatrix}$ & 0.931 & $\begin{smallmatrix}{0.99}\\{0.788}\end{smallmatrix}$ & \cmark & 0.623 \\[1.5mm]
        \bottomrule
    \end{tabular}
\end{table}
\begin{table}[ht]
    \centering
    \caption{Performance on the MM dataset.}
    \label{tab:mm}
    \begin{tabular}{c@{\hspace{0mm}}ccccccc}
        \toprule
        \multirow{2}{*}[-2.5mm]{\vrt{\scriptsize mode}} & \multirow{2}{*}[-2mm]{\vrt{\scriptsize tuning}} & \multicolumn{2}{c}{training R$^2$} & \multicolumn{2}{c}{testing R$^2$} & & mean \\
        \cmidrule(lr){3-4} \cmidrule(lr){5-6}
        & & median & $\begin{smallmatrix}\text{\scriptsize max}\\\text{\scriptsize min}\end{smallmatrix}$ & median & $\begin{smallmatrix}\text{\scriptsize max}\\\text{\scriptsize min}\end{smallmatrix}$ & vb & LCF \\
        \midrule
        - & - & 0.966 & $\begin{smallmatrix}{0.987}\\{0.954}\end{smallmatrix}$ & 0.96 & $\begin{smallmatrix}{0.983}\\{0.93}\end{smallmatrix}$ &   & 0 \\[1.5mm]
        U & B & 0.988 & $\begin{smallmatrix}{0.997}\\{0.973}\end{smallmatrix}$ & 0.985 & $\begin{smallmatrix}{0.995}\\{0.969}\end{smallmatrix}$ & \cmark & 0.763 \\[1.5mm]
        U & C & 0.988 & $\begin{smallmatrix}{0.996}\\{0.969}\end{smallmatrix}$ & 0.985 & $\begin{smallmatrix}{0.995}\\{0.943}\end{smallmatrix}$ & \cmark & 0.797 \\[1.5mm]
        S & B & 0.976 & $\begin{smallmatrix}{0.991}\\{0.967}\end{smallmatrix}$ & 0.973 & $\begin{smallmatrix}{0.986}\\{0.961}\end{smallmatrix}$ & \cmark & 0.559 \\[1.5mm]
        S & C & 0.974 & $\begin{smallmatrix}{0.997}\\{0.947}\end{smallmatrix}$ & 0.971 & $\begin{smallmatrix}{0.996}\\{0.935}\end{smallmatrix}$ & \cmark & 0.563 \\[1.5mm]
        \bottomrule
    \end{tabular}
\end{table}

\paragraph{Discussion}
In the summary results (see Table \ref{tab:summary}) we can see that the configurations using globally synchronized mode were either worse or of similar performance as the baseline.
We hypothesize that this is caused by the fact that the shared linear combinations are modified based on all individuals in the population of which some can have a good structure but equally some can have a bad structure.
In the end, good and bad individuals fight against each other which, in turn, makes the usefulness of the shared combinations doubtful.

We can also see that the (locally) synchronized mode does not perform as well as the unsynchronized mode.
We hypothesize that this is caused by the need to resolve conflicts when structural changes (i.e. mutation or crossover) happen (see Section \ref{sec:synced}).
However, after the synchronization the model can be much worse than its parent(s) before the crossover (or mutation).
Also, this synchronization takes additional time and slows the process down (though not very much).

Another result clearly visible from the summary results is that modifying the LCFs only with mutation makes almost no difference to the baseline.
This supports similar results on the toy problems.
On the other hand, every configuration using backpropagation (except GB which we already discussed) is better than the baseline more than once.
We see this as a clear indication that using the backpropagation technique to tune the LCF weights is a viable approach.

On the K11C dataset (see Table \ref{tab:k11c} and Figure \ref{fig:k11c}) we can see quite a strong result in favor of UB and UC configurations.
Again, this result confirms the benefit of the LCFs since the true relationship itself contains several constants that multiply or offset the variables -- precisely what the LCFs are capable of.

On the UB5D dataset (see Table \ref{tab:ub5d} and Figure \ref{fig:ub5d}), baseline was by far the best.
We hypothesize that this is caused by the fact that the true relationship is one big fraction but there is no division operator in the function set.
Because of this, the LCF-based configurations are, in principle, not able to find a structure that would enable them to find a good LCFs.
Therefore they are wasting time, compared to the baseline which does not attempt to do this and aims for approximating the relationship by ``brute force''.
Similar result, though not as strong, can be observed on the ASN dataset (see Table \ref{tab:asn} and Figure \ref{fig:asn}).
Since it is a real-world dataset, we don't know the true relationship and we cannot draw the same conclusions as for UB5D.

Worth noting is the fact that the selected presented configurations, especially UB and UC, performed very well on the datasets from reinforcement learning (RL) domain.
The first possible explanation could be that these configurations are good on RL value functions.
However, there are many possible RL problems and their value functions so such generalization cannot be made.
Second possible explanation is that a common feature of these functions -- they are all very smooth, without very sharp peaks, oscillations and noise -- is well suited for LCFs.
This is more viable explanation since the toy problems and the K11C dataset also share this feature, but it is still only a hypothesis.
The last possible explanation is that these datasets were easy because they are just two-dimensional.
The low dimensionality certainly plays an important role, though it is difficult to asses the measure of importance.
However, even if it was so, it would be an indication that LCFs are very good for problems of low dimensionality.

\section{Conclusions and Future work} \label{sec:conclusion}
In this article we presented a new type of leaf node for use in SR -- linear combination of feature variables -- which we then used in the baseline algorithm of MGGP.
We have presented two approaches to tuning their weights, one based on mutation and the other one based on error backpropagation technique.
We also presented three operation modes, one very flexible with no constraints, one enforcing creation of affine transformations of the feature space and one enforcing such a transformation on the whole population.

All sensible configurations of the proposed algorithm were tested on a set of benchmarks with the focus on showing the differences from the baseline algorithm.
The toy problems, designed specifically to test the ability of LCFs, showed that they are capable of handling existing feature space transformation.
The results on realistic problems have shown that configurations using globally synchronized mode are of no benefit or make the algorithm worse, and we provided a possible explanation.
On the other hand, two configurations stood out as clear improvement over the baseline.

The presented work is just a first glance at the possibilities.
The proposed approach was tested only on problems of low dimensionality.
A proper testing on high-dimensional problems is necessary.
Future research could also focus on tuning the parameters of the algorithm to find e.g. how much backpropagation tuning is suitable, or which update method fits best into this setting.
Another idea worth probing is extending this concept to other nodes than leaves, bringing similar tuning capabilities inside the trees.

\begin{acks}
    This work was supported by the Czech Science Foundation project Nr. 15-22731S with a student support from the Grant Agency of the Czech Technical University in Prague, grant No. SGS17\slash 093\slash OHK3\slash 1T\slash 13.
    Access to computing and storage facilities owned by parties and projects contributing to the National Grid Infrastructure MetaCentrum provided under the programme "Projects of Large Research, Development, and Innovations Infrastructures" (CESNET LM2015042), is greatly appreciated.
\end{acks}

\bibliographystyle{ACM-Reference-Format}
\bibliography{../../bibliography}

\end{document}